\documentclass{llncs}
\usepackage{graphicx}
\usepackage{authblk}    
\usepackage{hyperref}   
\usepackage{fancyhdr}   
\usepackage{booktabs}
\usepackage{multirow}
\usepackage{cleveref}
\usepackage{mathrsfs}  
\usepackage{array} 
\usepackage[a4paper, 
            left=3.18cm, 
            right=3.18cm, 
            top=2.54cm, 
            bottom=2.54cm]{geometry}
\usepackage{layout}
\usepackage{authblk}

\begin{document}
\title{X-Driver:Explainable  Autonomous Driving with Vision-Language Models}

\author{Wei Liu$^{\dagger1,2}$, Jiyuan Zhang$^{\dagger2}$, Binxiong Zheng$^2$, Yufeng Hu$^2$,Yingzhan Lin$^2$, Zengfeng Zeng$^{*2}$
\thanks{$^{\dagger}$ Equal Contribution, $^*$ For correspondence: {\tt\footnotesize gdfeng@126.com}.}
}


\institute{
  \textsuperscript{1}Harbin Institute of Technology, Shenzhen \quad
  \textsuperscript{2}Baidu Inc.}

\maketitle
\begin{abstract}

End-to-end autonomous driving has advanced significantly, offering benefits such as system simplicity and stronger driving performance in both open-loop and closed-loop settings than conventional pipelines. However, existing frameworks still suffer from low success rates in closed-loop evaluations, highlighting their limitations in real-world deployment. In this paper, we introduce X-Driver, a unified multi-modal large language models\cite{cui2024survey}(MLLMs) framework designed for closed-loop autonomous driving, leveraging Chain-of-Thought(CoT) and autoregressive modeling to enhance perception and decision-making. We validate X-Driver across multiple autonomous driving tasks using public benchmarks in CARLA simulation environment, including Bench2Drive\cite{jia2024bench2drive}. Our experimental results demonstrate superior closed-loop performance, surpassing the current state-of-the-art(SOTA) while improving the interpretability of driving decisions. These findings underscore the importance of structured reasoning in end-to-end driving and establish X-Driver as a strong baseline for future research in closed-loop autonomous driving.

\end{abstract}

\section{Introduction}

End-to-end autonomous driving has emerged as a promising paradigm that unifies perception, prediction, and decision-making within a single framework. While conventional modular pipelines offer interpretability, they often suffer from error propagation and struggle to generalize across diverse environments. Recent advancements in MLLMs with present new opportunities for leveraging diverse input modalities, such as natural language commands and visual data, to enhance autonomous decision-making.
Despite their potential, existing approaches face critical limitations. First, traditional autonomous driving systems rely on modular designs with fixed-format sensor inputs, making it challenging to effectively integrate multi-modal data. Second, current MLLM-based frameworks struggle with closed-loop evaluations, exhibiting hallucinations and a lack of robustness in real-world driving scenarios. These challenges highlight the need for explicit reasoning mechanisms to enhance both interpretability and reliability.
To address these issues, we introduce X-Driver, a unified framework that seamlessly integrates perception, reasoning, and decision-making for autonomous driving. X-Driver leverages autoregressive Transformers\cite{vaswani2017attention} for multi-modal fusion. Inspired by DriveCoT\cite{wang2024drivecot}, we employ CoT prompting, which structures the model’s reasoning process to enhance decision-making transparency and robustness. By using given prompts and chain-of-thought training templates, and applying them to tasks such as object detection, traffic detection, traffic sign detection, and lane detection, we perform step-by-step reasoning based on the data for final decision-making. Our experiments demonstrate that X-Driver is capable of explicitly providing the reasons behind its decisions. Extensive closed-loop evaluations demonstrate that this approach significantly improves performance over existing frameworks. 
Our contributions can be summarized as follows:
\begin{enumerate}
    \item MLLM and CoT Fusion: incorporating chain-of-thought reasoning into autonomous driving can further mitigate model hallucinations, significantly reducing the likelihood of decision-making errors.
    \item A unified closed-loop autonomous driving Framework: X-Driver supports multi-modal data inputs, moving beyond reliance on fixed-format sensor data. This flexibility improves the system’s generalization capabilities, allowing it to adapt more effectively to diverse and complex driving environments. And we already finished closed-loop experiments.
\end{enumerate}

\section{Related Works}

\subsection{Multi-modal Large Models for Autonomous Driving}
Multi-modal large models have demonstrated impressive capabilities in joint image-text understanding and reasoning. \cite{vemprala2024chatgpt} explores the integration of OpenAI's ChatGPT into robotics, proposing a framework that leverages natural language interaction and code synthesis to bridge human intent with robotic task execution. Transfusion\cite{zhou2024transfusion} exemplifies this paradigm by integrating autoregressive token prediction with diffusion-based image modeling\cite{ho2020denoising}. Another exciting work is LMDrive\cite{shao2024lmdrive}, which leverages large models to achieve closed-loop end-to-end autonomous driving with promising results. 
Vision-language pretraining frameworks, including Flamingo\cite{alayrac2022flamingo} and GPT-4V\cite{zhou2024gpt}, inspire the extension of multimodal learning into autonomous driving. These models provide valuable insights into fusing language, vision, and control modalities but lack dedicated designs for driving dynamics and safety-critical applications.
\subsection{Chain-of-Thought Reasoning in Driving}
DriveCoT\cite{wang2024drivecot} introduces Chain-of-Thought reasoning into end-to-end driving, providing interpretable rationales for control decisions. CoT-Drive\cite{liao2025cot} transfers the complex reasoning capabilities of LLMs to lightweight edge models through knowledge distillation and incorporates CoT prompting to enable efficient and accurate real-time motion forecasting for autonomous driving. Similar works include EMMA\cite{hwang2024emma}. \cite{zhao2025sce2drivex} integrates local scene videos and global bird’s-eye view maps to enhance 3D scene understanding and reasoning for autonomous driving, utilizing a CoT approach to bridge high-level semantic understanding and low-level motion control. Similar work includes \cite{luo2024pkrd}. While CoT improves explainability and trustworthiness, existing models are primarily language-based and lack seamless integration with continuous control and visual reasoning.

\section{Methodology}

The core of our system is LLaVA, a multi-modal model that aligns a vision encoder with a large language model, enabling it to understand images and perform dialogue and reasoning based on visual and textual inputs.

A typical reasoning process in MLLMs can be represented as follows:
\begin{equation}
     Y = \mathcal{F}(T,I)
\end{equation}
For an entire time series, the MLLM generates output commands in an autoregressive manner:
\begin{equation}
    P(Y|(T,I))=\prod_{t=1}^{T}P(y_t|y_1,\ldots,y_{t-1},(T,I))
\end{equation}
Where T represents the textual input, which unifies all prompts, navigation commands, and scene descriptions into a single coherent sequence to fully leverage the MLLM’s capability in language understanding and reasoning; I denotes the image input at each time step, including both the current and historical frames to provide temporal visual context; and $\mathcal{F}$ is our multimodal large language model, LLaVA, which has been supervised fine-tuned to better perform autonomous driving tasks. The model outputs Y, whose meaning varies across different stages of the chain-of-thought (CoT) process: in the perception stage, Y represents spatially grounded 3D objects formatted as (x, y, z, length, width, height), while in the prediction stage, Y denotes a sequence of waypoints, expressed as $(x_1, y_1), …, (x_k, y_k)$. In our system, we perform supervised fine-tuning on LLaVA to enhance its ability to handle autonomous driving tasks. 
\subsection{System Overview}
\begin{figure*}[t]
    \centering
    \includegraphics[width=\textwidth]{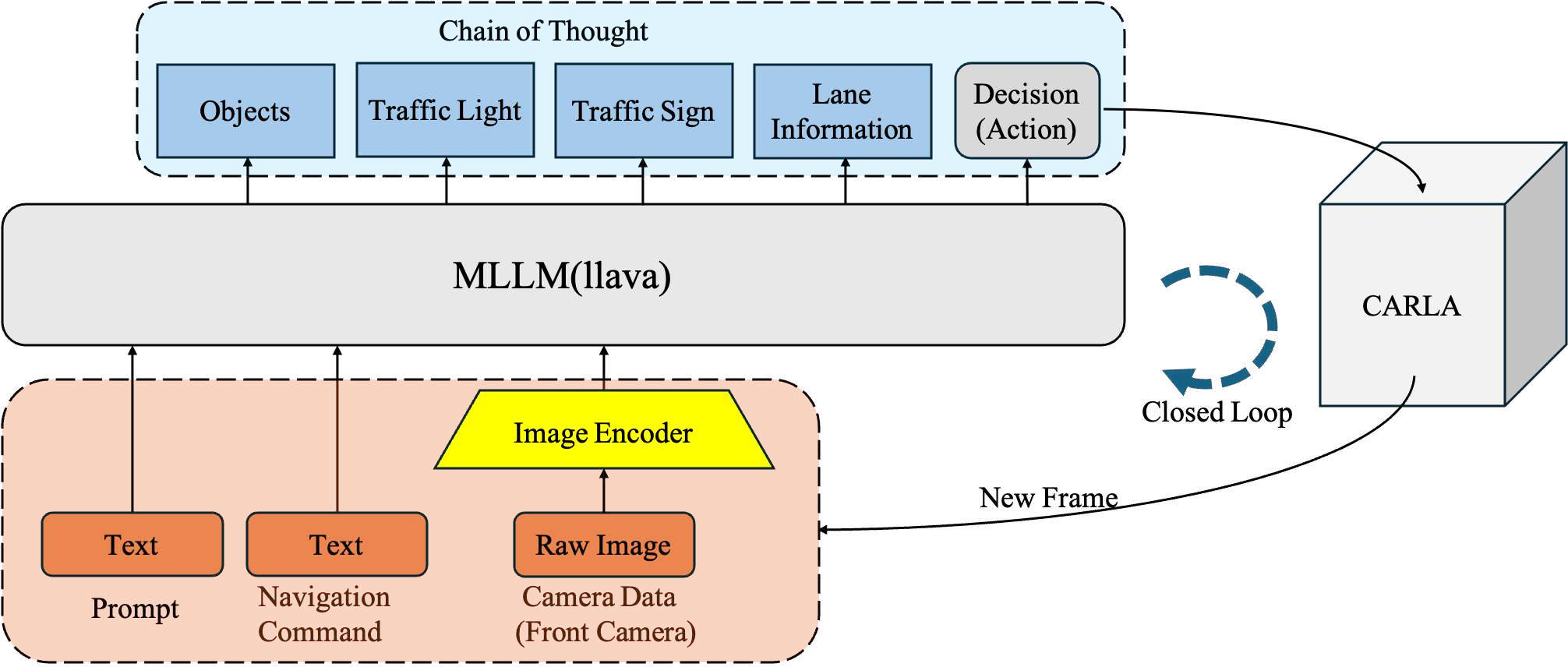} 
    \caption{\textbf{Overview of our system}}
    \label{fig:workflow}
    \vspace{-5mm}
\end{figure*}

As illustrated in \ref{fig:workflow}, our system leverages a MLLM(llava\cite{liu2024improved}) with an integrated CoT reasoning mechanism to enhance autonomous driving decision-making. The system processes multi-modal inputs, including image vectors from camera data and text vectors representing navigation commands and prompts. By employing CoT reasoning, the MLLM performs scene comprehension, navigation instruction interpretation, and traffic rule understanding to generate appropriate decisions to achieve exact driving actions. The system operates in a closed-loop manner, where executed actions influence the real-world environment, generating new perceptual data for continuous refinement. This approach improves decision explainability, enhances generalization in diverse scenarios, and strengthens the safety and robustness of end-to-end autonomous driving.
\subsection{MLLM and CoT Fusion}

In our MLLM, we enhance reasoning and decision-making capabilities in autonomous driving scenarios by leveraging Supervised Fine-Tuning (SFT) with high-quality CoT prompt data. Specifically, we incorporate meticulously crafted, step-by-step reasoning examples into the model inputs to encourage structured, logical thinking, rather than allowing the model to merely output isolated decisions.
Our CoT reasoning framework is built upon the model's comprehensive understanding across two critical dimensions: 
\begin{enumerate}
    \item It encompasses accurate perception and interpretation of complex 3D driving environments, including precise predictions of dynamic objects’ positions, velocities, and trajectories. These objects include pedestrians with varying walking speeds and intentions, cyclists who may suddenly change direction, and vehicles of different sizes (e.g., cars, trucks, motorcycles) that exhibit diverse driving behaviors such as acceleration, braking, or lane changes. Additionally, it ensures real-time obstacle identification, such as road debris, construction barriers, and stationary vehicles, along with precise spatial localization to maintain safe navigation.
    
    \item It involves deep comprehension of navigation instructions and adherence to traffic regulations, including recognizing traffic light statuses with differentiation between standard red, yellow, and green signals, as well as more complex variations such as flashing yellow lights or pedestrian-controlled signals. Furthermore, it requires interpreting traffic signs, which include regulatory signs (e.g., stop, yield, speed limits), warning signs (e.g., sharp turns, pedestrian crossings), and informational signs (e.g., highway exits, distance markers). Accurate lane detection and decision-making are also crucial, encompassing detection of lane boundaries in varying conditions (e.g., faded markings, occlusions by vehicles), differentiation between solid and dashed lines for legal lane changes, and identifying special lanes such as bus lanes, bicycle lanes, and turning-only lanes to ensure lawful and efficient driving.
\end{enumerate}
\begin{figure*}[h]
    \centering
    \includegraphics[width=\textwidth]{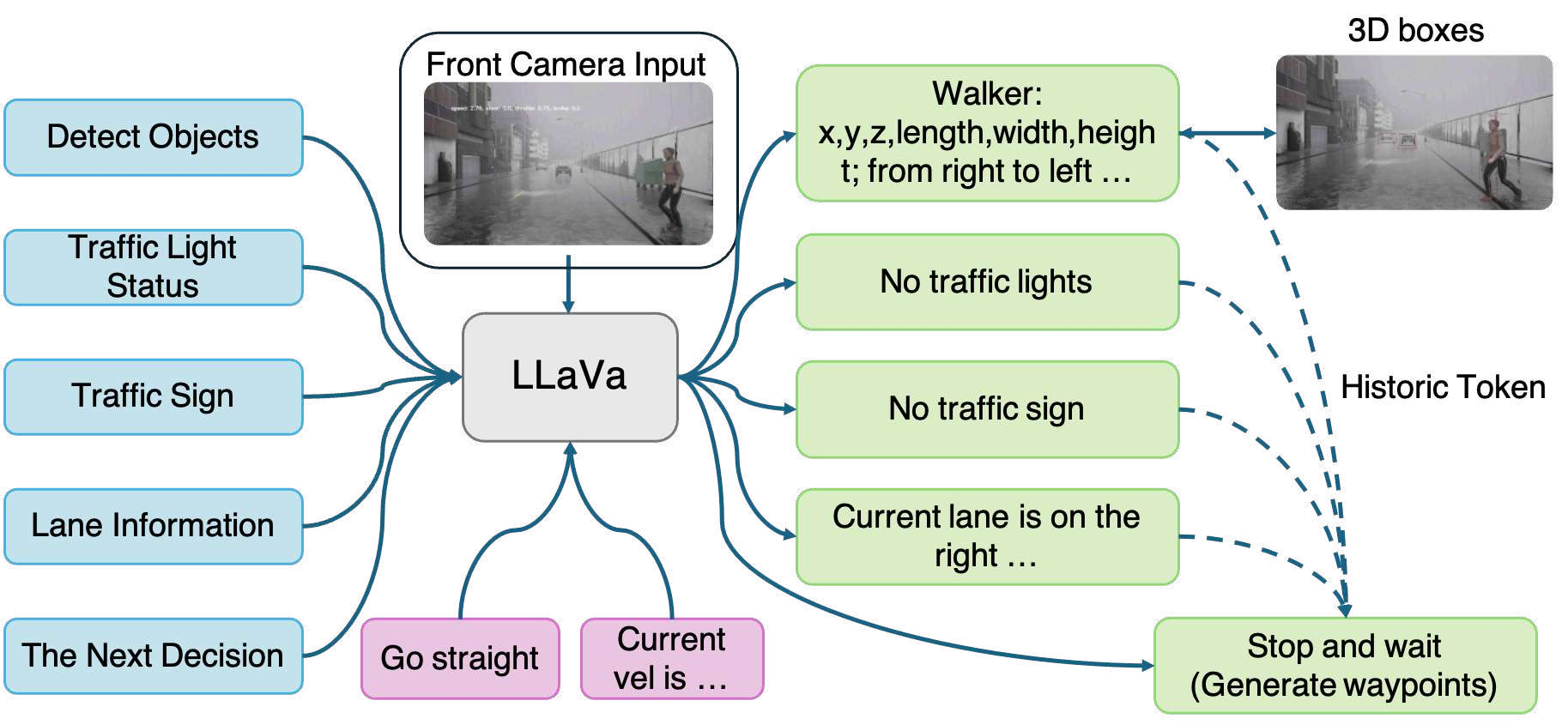} 
    \caption{\textbf{examples of Training Process}}
    \label{fig:train}
    \vspace{-5mm}
\end{figure*}
\noindent To ensure a higher performance and understanding of images, we avoid using discrete encoding methods like VQ-VAE\cite{van2017neural}, which may lead to information loss in scene representation. Instead, we adopt a continuous image encoding approach by first passing the raw image through a ViT encoder\cite{radford2021learning} to obtain a low-dimensional feature map. This strategy preserves richer scene information. For example, in our experiments, when a traffic light appears in the distant scene, the use of VQ-VAE encoding can result in the loss of crucial information about the traffic light. Conversely, utilizing VAE encoding effectively preserves this essential information. 

\begin{figure*}[t]
    \centering
    \includegraphics[width=0.8\textwidth]{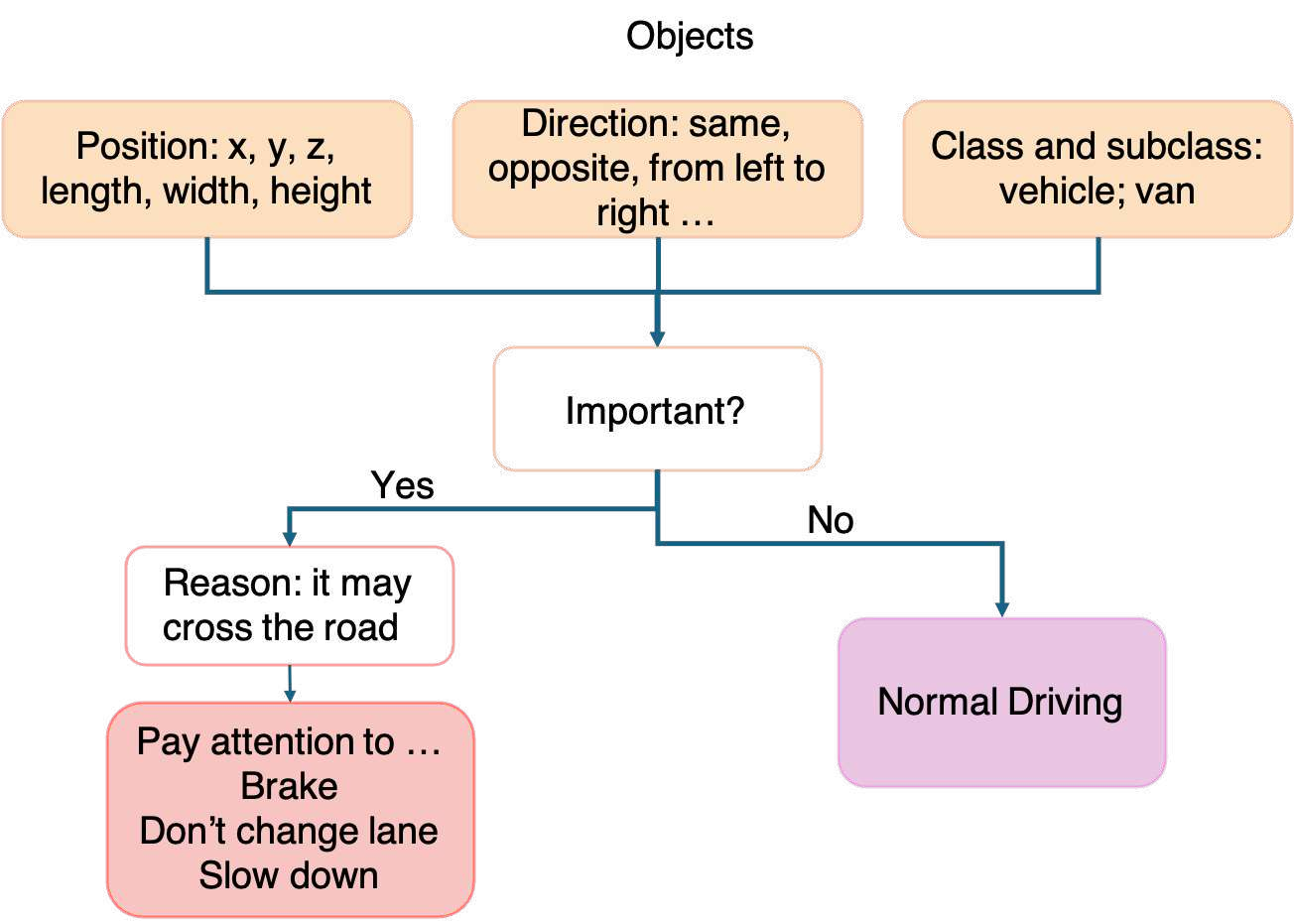} 
    \caption{\textbf{How CoT affects decision(Objects)}}
    \label{fig:detail}
    \vspace{-5mm}
\end{figure*}

As illustrated in \ref{fig:detail}, this is a detailed example of the SFT training process. Our CoT approach essentially decomposes the original task into four sub-tasks, such as object detection, traffic light status,Traffic sign and lane information. The model then integrates these intermediate results as historic tokens and current sensor inputs, to generate the final driving decision and predict the next waypoints. This structured reasoning enables more interpretable and controllable outputs in end-to-end autonomous driving.

\subsection{CoT Reasoning Process}
\noindent As depicted in \ref{fig:workflow}, our approach initiates with the generation of high-quality CoT training data that integrates camera inputs along with the current vehicle speed. Subsequently, the MLLM performs deep multi-modal fusion and analysis of these sensor inputs, systematically guided by CoT prompts to articulate a clear, step-by-step reasoning process.


Why is CoT reasoning so effective, and what role does it play in the inference process of large language models? Taking object detection within CoT as an example, as illustrated in \ref{fig:detail}, the model first determines an object’s position, motion direction, and category. It then analyzes whether this object warrants attention based on contextual cues. If attention is deemed necessary, the model provides a reasoning-based explanation for focusing on the object, and subsequently updates its final decision accordingly.

\noindent Table~\ref{tab:instruction} illustrates typical driving scenarios encountered in autonomous driving, paired with detailed prompts that guide the MLLM through explicit, logical decision-making steps. These prompts direct the model to incorporate comprehensive situational awareness and adhere strictly to safety guidelines and traffic rules, significantly enhancing driving safety and reducing erroneous model behavior.
\begin{table}[h]
    \centering
    \caption{Examples of Driving Scenario Prompts and Corresponding Model Instructions}
    \begin{tabular}{m{4.6cm}|m{9.6cm}}
        \hline
        \multicolumn{1}{c|}{\textbf{Scenario}} & \multicolumn{1}{c}{\textbf{Model Instruction}} \\
        \hline
        
        Red light (approaching at speed) & Slow down to a complete stop and wait for the light to turn green. \\
        \hline
        Red light (stationary) & Stop and wait for the light to turn green.\\
        \hline
        Green light & With the green light, [Pay attention to xxx.], safely [turn left] and cross the intersection.\\
        \hline
        Pedestrian crossing ahead & Stop and wait for pedestrians crossing the road ahead.\\
        \hline
        Vehicle ahead within 20m (same lane) & [There is a vehicle ahead.] Maintain a safe following distance.\\
        \hline
        Turn left/right/straight (junction) & [Turn left] and [Pay attention to xxx.] \\
        \hline
        Vehicle ahead changing lanes & [Pay attention to the vehicle ahead changing lanes.] Reduce speed and maintain a safe following distance.\\
        \hline
        Ego-vehicle lane change & 1. Lane change triggered by navigation command.
        \newline 2. Risk management: Ensure safe lane change conditions, considering speed and position of surrounding vehicles. 
        \newline a. Utilize adjacent lanes in the same direction. 
        \newline b. Consider lanes in the opposite direction if necessary\\
        \hline
        Approaching exit ramp & [Approaching exit ramp, reduce speed, and enhance environment observation.]\\
        \hline
        Default driving & Normal driving behavior, maintaining vigilance and adhering to general traffic regulations.\\
        \hline
    \end{tabular}
    
    \label{tab:instruction}
\end{table}

\subsection{A Closed-loop Autonomous Driving Framework}

As demonstrated by the flowchart in \ref{fig:workflow}, our X-Driver framework systematically generates driving decisions through comprehensive understanding and reasoning based on navigation instructions and image data. This approach not only outputs actionable decisions but also provides explicit reasoning behind each decision, ensuring interpretability and robustness in driving actions within the CARLA simulation environment. Specifically, the MLLM generates real-time driving commands and corresponding waypoint predictions based on current sensor inputs and vehicle speed. Subsequently, the ego-vehicle dynamically adjusts its movements based on these waypoints and decision commands, thus maintaining a closed-loop control mechanism. Our methodology achieves SOTA performance on the Bench2Drive dataset, demonstrating significant advancements in autonomous driving capabilities.

\section{Experiments}

\subsection{Benchmarks and Datasets}
Bench2Drive: A simulation benchmark with 2 million+ frames, evaluating closed-loop driving performance under varying conditions (urban, highway, adverse weather).

\subsection{Open-loop Experiments}
In the CoT section, we compared the recognition performance at different stages of CoT. 
\subsubsection{3D object recognition}
In the 3D object recognition experiment, we tested the accuracy of our 3D bounding box recognition by calculating the IoU. The quantitative results of the experiment are shown in Table~\ref{tab:object recognition}.
\begin{table}[ht]
\centering

\small
\caption{\footnotesize{Results of Object Recognition}}
\label{tab:object recognition}
    \begin{tabular}{ccccccc}
    \hline
    \multirow{2}{*}{Dataset}& \multicolumn{6}{c}{3D IoU $\geq 0.5$}\\
    \cmidrule(lr){2-7}
    &IoU(sample) & IoU(box) & Precision & Recall & pred(total) & gt(total) \\
    bench2drive & 0.724 &0.806&0.678&0.673&3165&3191\\

    \hline
    \end{tabular}
\end{table}

In this table, IoU(sample) and IoU(box) measure the overlap between predicted and ground truth boxes, with the former being sample-level evaluation and the latter being box-level evaluation. Precision indicates the proportion of true positives among the predicted positives, and Recall represents the proportion of true positives correctly predicted. Pred(total) is the number of predicted objects by the model, and gt(total) is the number of ground truth objects in the dataset.  

From this table, we can see that my object recognition method achieved a remarkably high accuracy.

\subsubsection{Waypoint Accuracy}
To evaluate the accuracy of our model in future trajectory prediction, we conducted open-loop experiments on waypoint errors. The experimental results are shown in Table~\ref{tab:waypoint}

Where ADE (Average Displacement Error) and FDE (Final Displacement Error) are two key evaluation metrics used to measure the deviation between predicted and ground truth trajectories.

The experimental results show that our large model achieves prediction accuracy of less than 1\% for future trajectories, demonstrating excellent performance.
\begin{table}[ht]
\centering
\label{tab:waypoint}
\small
\caption{\footnotesize{Results of Waypoints Accuracy}}
\label{tab:waypoint}
    \begin{tabular}{cccccc}
    \hline
    \multirow{2}{*}{Dataset}& \multicolumn{4}{c}{ADE} & \multirow{2}{1.2cm}{\centering FDE}\\
    \cmidrule(lr){2-5}
    &0.5s & 1s & 2s & 3s & \\
    bench2drive & 0.679 &0.837&1.128&1.488&2.472\\

    \hline
    \end{tabular}
\end{table}

\subsection{Closed-loop Experiments}
We evaluated our approach in a closed-loop setting using the CARLA simulation environment, focusing on Driving Score and Success Rate as key performance metrics. Driving Score assesses overall driving quality by considering factors such as route adherence, speed control, and traffic rule compliance, while Success Rate measures the percentage of successfully completed driving tasks, ensuring the vehicle reaches its destination without collisions or major infractions.

\subsubsection{Qualitative Study}
\begin{figure*}[h]
    \centering
    \includegraphics[width=\textwidth]{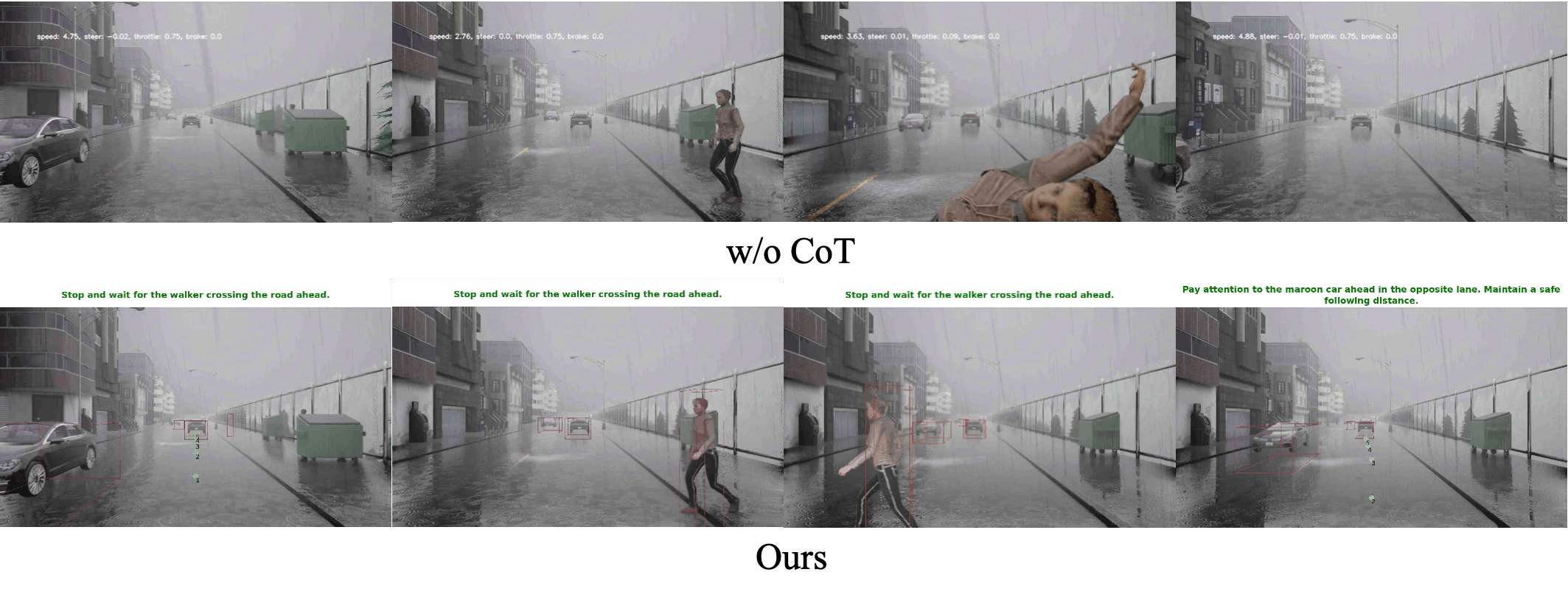} 
    \caption{Comparison of Utilizing CoT or not}
    \label{fig:compare}
    \vspace{-5mm}
\end{figure*}

In the closed-loop ablation experiments illustrated in \ref{fig:compare}, the clear advantage of the CoT approach is evident. Specifically, the CoT version successfully identifies pedestrians crossing the road and promptly initiates braking to avoid collisions. In contrast, the auxiliary-task-based version fails to detect pedestrians, resulting in collisions.

\subsubsection{Quantitative Study}
We evaluated our framework against the current SOTA UniAD\cite{hu2023planning} on the Bench2Drive dataset with 500K and 2.2M samples. The results depicted in Table~\ref{tab:result} demonstrate that our approach consistently outperforms the SOTA across multiple datasets on key metrics, further confirming that incorporating CoT reasoning can significantly enhance the decision-making accuracy of MLLMs.
\begin{table}[ht]
\centering
\large
\caption{\footnotesize{Results of Closed-loop Experiments}}
\label{tab:result}
    \resizebox{\linewidth}{!}{
    \begin{tabular}{m{4.8cm}cccc}
    \hline
    & \multicolumn{2}{c}{bench2drive220} &\multicolumn{2}{c}{bench2drive50} \\ 
    \cmidrule(lr){2-3} \cmidrule(lr){4-5}
    & Driving Score$\uparrow$ & Success Rate(\%)$\uparrow$ & Driving Score$\uparrow$ & Success Rate(\%)$\uparrow$\\
    \hline

    UniAD(2 million frames bench2drive) & 45.9 & 17.5 & 49.1 & 22.0 \\
    Ours(w/o CoT: 3 million frames) & 49.1 & 15.2 & 52.5 & 19.8 \\
    Ours(CoT: 3 million frames, auxiliary task version, split into 4 subtasks) & 51.7 & 18.1 & 57.8 & 24.0 \\
    \hline
    \end{tabular}
    }
\end{table}

\section{Conclusion}
Overall, we propose a unified end-to-end autonomous driving framework that leverages a MLLM for decision-making while integrating CoT reasoning to enhance interpretability. By explicitly modeling the reasoning process, our approach effectively mitigates hallucinations and decision errors, leading to more consistent, reliable, and human-aligned driving behaviors. Furthermore, our framework significantly improves closed-loop driving performance in autonomous systems by enhancing the model’s ability to comprehend complex driving scenarios, adhere to traffic regulations, and make informed, context-aware decisions in real time.

\newpage
\bibliographystyle{splncs04}

\end{document}